\documentclass[smallextended]{svjour3}       
\usepackage[bottom=4cm, right=4cm, left=4cm, top=4cm]{geometry}
\usepackage{amsfonts}
\usepackage{latexsym}
\usepackage{amssymb}
\usepackage{amsmath}
\usepackage[mathscr]{eucal}
\usepackage{graphicx}
\usepackage{hyperref}
\usepackage{caption}
\usepackage{subcaption}

\usepackage{algorithm}
\usepackage[noend]{algpseudocode}

\renewcommand{\qed}{\hfill{\ \ \rule{2mm}{2mm}} \vspace{0.2in}}

\newcommand{\ind}{1\hspace{-2.3mm}{1}}

\begin{document}

\title{Dissimilar Batch Decompositions of Random Datasets}
\titlerunning{Dissimilarity in Random Datasets}

\author{ \textbf{Ghurumuruhan Ganesan}
\thanks{E-Mail: \texttt{gganesan82@gmail.com} } \\
\ \\
IISER Bhopal}


\date{}
\maketitle

\begin{abstract}
For better learning, large datasets are often split into small batches and fed sequentially to the predictive model. In this paper, we study such batch decompositions from a probabilistic perspective. We assume that data points (possibly corrupted) are drawn independently from a given space and define a concept of similarity between two data points. We then consider decompositions that restrict the amount of similarity within each batch and obtain high probability bounds for the minimum size. We demonstrate an inherent tradeoff between relaxing the similarity constraint and the overall size and also use martingale methods to obtain bounds for the maximum size of  data subsets with a given similarity. 

\vspace{0.1in} \noindent \textbf{Key words:} Random Datasets; Corrupted Entries; Dissimilar Batch Decompositions; Martingale method.

\vspace{0.1in} \noindent \textbf{AMS 2000 Subject Classification:} Primary: 60K35, 60J10;
\end{abstract}

\bigskip

\renewcommand{\theequation}{\thesection.\arabic{equation}}
\setcounter{equation}{0}
\section{Introduction} \label{intro}
Large datasets fed into predictive models are often split into small batches in order to reduce computational operations and memory usage. Another implicit advantage is that this also improves the performance of the gradient descent algorithms running in the background~\cite{kuhn}: if each batch contains sufficiently diverse data, then we expect that the model  learns the patterns better and thereby ensuring that the overall performance in terms of accuracy is improved.

In this paper, we study batch decompositions of datasets from a probabilistic perspective. We assume that the data points are drawn independently from a given space and are also possibly subject to corruption which happens, for example, due to data unavailability or simply human error.  We use martingale methods along with segmentation techniques used in random geometric graphs~\cite{gupta}~\cite{penrose} to obtain high probability bounds for the minimum size of constrained decompositions and highlight the tradeoff between relaxing the similarity parameter and the resulting decomposition size.

We have a couple of remarks regarding the above paragraph:\\
\underline{\emph{Remark 1}}: In practice, corrupted entries are filled by using some form of imputation before being fed to the predictive model~\cite{mike}. Throughout, however, we consider the original dataset \emph{before} any modifications are performed and describe and analyze batch decompositions that reduce similarity within each batch.\\\\
\underline{\emph{Remark 2}}: Our motivation  behind the dissimilar batch decomposition is this: Consider for example a data set consisting of~\(1000\) images that contains~\(100\) ``basic" images and the rest being rotation and scaling of the basic set of images. This augmentation is often done to increase the size of sparse datasets~\cite{xu}. If we naively split the dataset into~\(100\) batches, each containing~\(10\) images, it may be possible that each batch contains only a few basic images (in the extreme case, exactly one basic image per batch). Feeding this suboptimal batchwise decomposition to the predictive model like artificial neural network  (ANN), might not be efficient since the network learns very little from any single batch. This is a major reason we  advocate dissimilar batch decompositions in this paper.

Finally, we also obtain bounds for the maximum size of data subsets with a given similarity. Similarity or redundancy in datasets is an important object of study from both theoretical and application perspectives. While redundancy may be useful when data is sparsely available, it is clearly undesirable from a storage point of view~\cite{kuhn}. Feature domain redundancy in datasets has been well-studied and there are well-known statistical methodologies to extract optimized subsets of features that result in low performance degradation (see chapter~\(19\) of~\cite{kuhn}).

Recently, in~\cite{ganesan} we have studied \emph{index} redundancy in the context of data undersampling for class imbalanced datasets. Class imbalance is very common in real world datasets prepared for classification and this adversely affects the performance of the predictive models (like for e.g.~\(k-\)Nearest Neighbour~(\(kNN\)), Random Forest, Logistic Regression or Neural Networks)~\cite{fern}.  Many ad hoc undersampling methods are known (like random undersampling, near miss, condensed neighbour etc.)~\cite{sui} and in~\cite{ganesan}, we use random graph techniques to propose and analyze a neighbourhood domination based undersampling methodology.

In this paper, we continue the study of index redundancy in datasets from a probabilistic perspective. We use martingale and iteration methods to obtain maximum size of data subsets with a given redundancy, i.e., similarity.

In the following Section, we state and prove our main result regarding the minimum size of batch decompositions that restrict the number of similar data points within each batch. We also use iterative techniques to estimate maximum size of data subsets with a given similarity and illustrate how the continuous and categorical parts of the data affect the overall size.


For convenience, we have collected the commonly used symbols and their representation, in  Table~\(1\) below.
\begin{table}[h!]
  \begin{center}
    \caption{Notation.}
    \label{tab:table1}
    \begin{tabular}{l|l} 
      \textbf{Symbol} & \textbf{Meaning}\\
      \hline
      \(W_j = (X_j,Y_j)\) & \(j^{th}\)~\text{data point }\\
      \(X_j\) & \text{ continuous part of} \(W_j\) \\
      \(Y_j\) & \text{ categorical part of} \(W_j\)\\
      \(p_0\) & \text{ corruption probability of}~\(X_j\)\\
      \(d\) & \text{ dimension of } \(X_j\)\\
      \(f\)  & \text{ density of } \(X_j\)\\
      \(S_0\) & \text{ square where}~\(f\) is positive\\
      \(\epsilon_{low}\) & \text{ lower bound of}~\(f\) in~\(S_0\)\\
      \(\epsilon_{up}\) & \text{ absolute upper bound for}~\(f\)\\
      \(p_{low}\) & \text{ minimum probability of occurrence of a categorical symbol}\\
      \(p_{up}\) & \text{ maximum probability of occurrence of a categorical symbol}\\
      \(N(j)\) & \text{ similarity set  of}~\(W_j\)\\
      \(\pi_d\) & \text{ volume of unit ball  in}~\(d\) dimensions\\
    \end{tabular}
  \end{center}
\end{table}


\setcounter{equation}{0}
\renewcommand\theequation{\thesection.\arabic{equation}}
\section{Random Datasets}\label{sec_sim}
For integer~\(d \geq 1\) let~\(\{X_j\}_{1 \leq j \leq n}\) be independent and identically distributed (i.i.d.) random vectors in~\(\mathbb{R}^d\) with common density~\(f.\) The integer~\(d\) does not depend on~\(n\) and we refer to~\(\{1,2,\ldots,n\}\) as the set of \emph{indices}. Let~\({\cal Y} = {\cal Y}(n)\) be any finite set and let~\(\{Y_j\}_{1 \leq j \leq n}\) be i.i.d.\ random elements in~\({\cal Y}\) (independent of~\(\{X_j\}\)) with distribution~\(p(y):= \mathbb{P}(Y_j=y).\) Finally, let~\( \delta_j, 1 \leq j \leq n\) be the i.i.d.\ Bernoulli random variables (that are independent of both~\(\{X_j\}\) and~\(\{Y_j\}\)) with distribution
\begin{equation}\label{del_dist}
\mathbb{P}(\delta_j = 1) = p_0 = 1-\mathbb{P}(\delta_j = 0).
\end{equation}
We define~\(W_j := (\delta_jX_j,Y_j)\) to be the~\(j^{th}\) data point with the understanding that~\(W_j\) is uncorrupted if~\(\delta_j=1\) and corrupted otherwise. We denote~\(\{W_j\}_{1 \leq j \leq n}\) to be the \emph{dataset}.

We say that data point~\(W_i\) is similar to~\(W_j\) if
\begin{equation}\label{sim_cond_two}
Y_i = Y_j \text{ and } \text{ either } \delta_i \delta_j = 1 \text{ and } d(X_i,X_j) < r_n  \text{ or } \delta_i\delta_j=0
\end{equation}
where~\(d(a,b)\) is the Euclidean distance between~\(a\) and~\(b.\) In words, if both~\(W_i\) and~\(W_j\) are uncorrupted, then~\(W_i\) is similar to~\(W_j\) if and only if  their categorical parts agree and the continuous parts are within distance~\(r_n\) of each other. On the other hand, if either~\(W_i\) or~\(W_j\) is corrupted (and hence comparison between the continuous parts is not possible), then  we declare~\(W_i\) to be similar to~\(W_j\) if their categorical parts are identical.


We have a few remarks:\\
\underline{\emph{Remark 3}}: For technical simplicity we have assumed that the dimension~\(d\) is a constant not depending on the size~\(n\) of the dataset and our bounds obtained below are also computable for the case when~\(d\) varies with~\(n.\) However, we have allowed the size of the categorical space to possibly  depend on~\(n,\) as is the case in many real world datasets particularly related to genetics, where the number of features far exceeds the number of samples~\cite{kuhn}~\cite{sharbaf}~\cite{zawbaa}.\\\\
\underline{\emph{Remark 4}}: For analytical convenience, we assume that only the continuous part of the data point is corrupted and our analysis below holds if the categorical part is itself a vector and at least one entry in a fixed position remains uncorrupted with probability one. We also assume that the whole vector~\(X_j\) is corrupted even if a single entry is corrupted. A similar analysis as below would hold for analysis of partially corrupted entries provided, we consider densities of all possible subsets of the~\(d\) entries and impose corresponding upper and lower bounds for each such density.\\

Our first step in the study of dataset batch decomposition is to obtain bounds for the number of data points similar to a given data point.
To that end, we define
\begin{equation}\label{n_sim_def}
N(v) := \{u: W_u \text{ is similar to }W_v\}
\end{equation}
to be the set of all indices of data points similar to~\(W_v\) and have the following Lemma. Throughout constants do not depend on~\(n\) and~\(p_{up} := \max_{y \in {\cal Y}}p(y)\) is the maximum probability of occurrence of a categorical symbol.
\begin{lemma}\label{lem_sim}
Suppose~\(r_n\) is bounded and there is a square~\(S_0\) of constant side length and constants~\(\epsilon_{low}\) and~\(\epsilon_{up}\) such that
\begin{equation}\label{daldamma}
0 < \epsilon_{low} \leq \min_{x \in S_0} f(x) \leq \max_{x \in \mathbb{R}^d} f(x) \leq \epsilon_{up} < \infty.
\end{equation}
There are constants~\(\gamma_1,\gamma_2 > 0\) such that
\begin{equation}\label{rasathi_ax}
\mathbb{P}(\gamma_1\Delta \leq \max_{v}\#N(v) \leq \gamma_2\Delta) \geq 1-2ne^{-\gamma_1\Lambda}
\end{equation}
where
\begin{equation}\label{delta_def}
\Delta := np_{up}\max(r_n^{d}(1-p_0), p_0) \text{ and }
\Lambda := \left\{
\begin{array}{ll}
np_{up}\min(r_n^{d}(1-p_0),p_0), & p_0  > 0 \\
np_{up}r_n^{d}, & p_0= 0.
\end{array}
\right.
\end{equation}
\end{lemma}
We have a few remarks:\\
\emph{\underline{Remark 5}}: The condition~(\ref{daldamma}) above does \emph{not} require that the support~\(S_f\) of the density~\(f\) is bounded and so the upper bound in~(\ref{daldamma}) is over the whole space~\(\mathbb{R}^d.\)  In other words, the continuous part of the data point could be unbounded (thereby allowing for the normality assumption frequently used while evaluating/scaling datasets~\cite{kuhn}). The square~\(S_0\) in~(\ref{daldamma}) could represent a high density region where a lot of datapoints cluster.\\\\
\underline{\emph{Remark 6}}: The constants~\(\gamma_1\) and~\(\gamma_2\) depend on the quantities~\(\epsilon_{low},\epsilon_{up}\) and the side length~\(a\) of the square~\(S_0.\) In fact from the proof below, we see that~\(\gamma_1 = C_1\min\left(1,\epsilon_{low}a^d\right)\) and~\(\gamma_2 = C_2 \epsilon_{up}\) for some absolute constants~\(C_1,C_2 > 0.\)\\\\
\underline{\emph{Remark 7}}: The parameter~\(\Delta\) essentially represents the growth of the largest size of a similarity set and plays a crucial role in analyzing batch decompositions later. To ensure that~(\ref{rasathi_ax}) holds with high probability, i.e., with probability converging to one as~\(n \rightarrow \infty,\) the term~\(\Lambda\) must grow at least of the order of~\(\log{n}.\) From~(\ref{delta_def}), this implies that we must choose the tolerance parameter~\(r_n\) to be at least of the order of~\(\left(\frac{\log{n}}{n}\right)^{1/d}.\) We return to this aspect later in the example following our next result concerning the size of batch decompositions.\\

Throughout, we use the following results regarding the deviation estimates of sums of independent Bernoulli random variables and the Lov\'asz Local Lemma, which we state together as a separate Lemma for convenience.
\begin{lemma}\label{lemmax}
\((i)\) Let~\(\{U_j\}_{1 \leq j \leq r}\) be independent Bernoulli random variables satisfying\\\(\mathbb{P}(U_j = 1) = 1-\mathbb{P}(U_j = 0) > 0.\) If~\(V_r := \sum_{j=1}^{r} U_j, \theta_r := \mathbb{E}V_r\) and~\(0 < \gamma \leq \frac{1}{2},\) then
\begin{equation}\label{conc_est_f}
\mathbb{P}\left(\left|V_r - \theta_r\right| \geq \theta_r \gamma \right) \leq 2\exp\left(-\frac{\gamma^2}{4}\theta_r\right)
\end{equation}
for all \(r \geq 1.\)\\
\((ii)\) Let~\(A_1,\ldots,A_t\) be events in an arbitrary probability space. Let~\(\Gamma \) be the dependency graph for the events~\(\{A_i\},\) with vertex set~\(\{1,2,\ldots,t\}\) and edge set~\({\cal E};\) i.e. assume that each~\(A_i\) is independent of the family of events~\(A_j, (i,j) \notin {\cal E}.\) If there are reals~\(0 \leq z_i < 1\) such that~\(\mathbb{P}(A_i) \leq z_i \prod_{(i,j)\in {\cal E}} (1-z_j)\) for each~\(i,\) then~\[\mathbb{P}\left(\bigcap_{i} A^c_i\right) \geq \prod_{1 \leq i \leq t} (1-z_i) > 0.\]
\end{lemma}
For proofs of Lemma~\ref{lemmax}\((i)\) and~\((ii),\) we refer respectively to Corollary~\(A.1.14,\) pp.~\(312\)  and Lemma~\(5.1.1,\) pp.~\(64\) of~\cite{alon}.

\emph{Proof of Lemma~\ref{lem_sim}}: We assume below that~\(p_0 >0\) and an analogous analysis holds for the case~\(p_0=0.\) To estimate the size of the similarity set~\(N(v),\) we begin with some preliminary computations. Let~\(y_0 \in {\cal Y}\) be an element of the categorical space whose conditional probability of occurrence~\(p(y_0) = p_{up}\) is the largest and set~\({\cal Z}(y_0) := \{j : Y_j=y_0\}\) to be the set of indices of all corrupted data points whose categorical part is~\(y_0.\) We see that~\(\#{\cal Z}(y_0)\) is Binomially distributed with parameters~\(n-1\) and~\(p_{up}p_0\) and so defining~\(E_{cat}(y_0) := \left\{\#{\cal Z}(y_0) \geq \frac{np_{up}p_0}{2} \right\}\) and using~(\ref{conc_est_f}) we get
\begin{equation}\label{zy_est_two}
\mathbb{P}\left(E_{cat}(y_0)\right) \geq 1- \exp\left(-C_1np_{up}p_0\right),
\end{equation}
for some constant~\(C >0.\)

Next, let~\(S_{1} \subset S_0\) be any~\(\frac{r_n}{\sqrt{4d}} \times \frac{r_n}{\sqrt{4d}}\) square contained within~\(S_0\) so that any two points of~\(S_1\) are within a distance~\(r_n\) from each other. The continuous part~\(X_k\) of the~\(k^{th}\) data point is uncorrupted and present in~\(S_1\) with probability~\[(1-p_0)\int_{S_1} f \geq 2C_2r_n^{d}(1-p_0)\] for some constant~\(C_2 > 0,\) by the lower bound in~(\ref{daldamma}). The categorical part of each such data point is~\(y_0\) with probability~\(p_{up}\) and so if~\(I_1\) is the number of uncorrupted data points whose continuous part lies in~\(S_1\) and whose categorical part is~\(y_0,\) then~\(I_1\) is stochastically dominated from below by a Binomial random variable with parameters~\(n\) and~\(2C_2r_n^{d}(1-p_0)p_{up}\) and so defining~\[E_{cont}(y_0) := \{I_1 \geq C_2nr_n^{d}(1-p_0)p_{up}\}\] we get from~(\ref{conc_est_f}) that
\begin{equation}\label{i_one_est}
\mathbb{P}\left(E_{cont}(y_0)\right)\geq 1-e^{-C_3nr_n^{d}(1-p_0)p_{up}}
\end{equation}
for some constant~\(C_3 > 0.\)

Suppose that~\(E_{tot}:= E_{cat}(y_0) \cap E_{cont}(y_0)\) occurs, which, from~(\ref{zy_est_two}),~(\ref{i_one_est}) and the union bound, happens with probability
\begin{equation}\label{e_tot_est}
\mathbb{P}(E_{tot}) \geq 1-e^{-C_1np_{up}p_0} - 2e^{-C_3nr_n^d(1-p_0)p_{up}} \geq 1-3e^{-C_4\Lambda}
\end{equation}
for some constant~\(C_4 > 0,\) where~\(\Lambda\) is as in Theorem statement. By definition this implies that there is a corrupted data point~\(W_{v_0}\) whose categorical part is~\(y_0\) and moreover~\(W_{v_0}\) is similar to at least~\(\frac{np_{up}p_0}{2}\) other corrupted data points. Similarly there is an uncorrupted data point~\(W_{v_1}\) with categorical part~\(y_0\) and whose continuous part lies in~\(S_1\) that is similar to at least~\(C_0nr_n^{d}p_{up}(1-p_0)-1\) other data points. This obtains the lower bound in~(\ref{rasathi_ax}).

To determine the upper deviation bounds in~(\ref{rasathi_ax}) and~(\ref{rasathi}), we first obtain a uniform upper bound for the size of the similarity sets  and then perform a random assignment strategy. We again precede with some preliminary calculations. Defining the sets~\(N_1(v) := \{a : \delta_a=0 \text{ and }Y_a = Y_v\}\) and \[N_2(v) := \{a : \delta_a = 1 \text{ and } d(X_a,X_v) < r_n \text{ and } Y_a=Y_v\},\] we see that~\(N(v) \subset N_1(v) \cup N_2(v)\) and so it suffices to estimate the size of the latter two sets. We begin with~\(N_1(v).\) Given~\(Y_v=y,\) the probability that~\(W_j, j \neq v\) is corrupted and the categorical part of~\(W_j\) also equals~\(y\) is~\(p(y)p_0 \leq p_{up}p_0.\) Therefore irrespective of~\(y,\) the term~\(\#N_1(v)\) is stochastically dominated from above by a Binomial random variable with parameters~\(n\) and~\(p_{up}p_0.\) Using the deviation estimate~(\ref{conc_est_f}), we then get
\begin{equation}\label{n_one}
\mathbb{P}\left(\#N_1(v) \leq 2np_{up}p_0\right) \geq 1-e^{-D_1np_{up}p_0}
\end{equation}
for some constant~\(D_1 > 0.\)

Next given~\(X_v=x,Y_v=y\) the number of uncorrupted data points whose continuous part lies in~\(B(x,r_n)\) and whose categorical part is~\(y,\) is Binomially distributed with parameters~\(n-1\) and~\[(1-p_0)p(y) \int_{B(x,r_n)} f \leq \epsilon_{up}\pi_d r_n^{d}p_{up}(1-p_0),\] by the upper bound for the density in~(\ref{daldamma}). Therefore choosing the constant~\(D_1 >0\) smaller if necessary and using~(\ref{conc_est_f}), we get~\[\mathbb{P}\left(\#N_2(v) \leq D_2 nr_n^{d}p_{up}(1-p_0)\right) \geq 1- e^{-D_1nr_n^{d}p_{up}(1-p_0)}\]
for some constant~\(D_2 > 0.\) Combining with~(\ref{n_one}) and using the union bound and the fact that~\(\#N(v) \leq \#N_1(v) + \#N_2(v),\) we get that
\begin{align}\label{n_tot}
\mathbb{P}\left(\#N(v) \leq D_3\Delta\right) &\geq 1-e^{-D_1np_{up}p_0}-e^{-D_1nr_n^{d}p_{up}(1-p_0)} \nonumber\\
&\geq 1-2e^{-D_1\Lambda}
\end{align}
for some constant~\(D_3 > 0,\) where~\(\Delta\) and~\(\Lambda\) are as in Theorem statement. Defining \[E_{sq} := \bigcap_{1 \leq  v\leq n} \{\#N(v) \leq D_3\Delta\},\] we get from the union bound that
\begin{equation}\label{sq_est}
\mathbb{P}(E_{sq}) \geq 1- 2n \cdot \exp\left(-D_1\Lambda\right).
\end{equation}
If~\(E_{sq}\) occurs, then each data point is similar to at most~\(D_1\Delta\) other data points and this proves the upper bound in~(\ref{rasathi_ax}) and therefore completes the proof of Lemma~\ref{lem_sim}.~\(\qed\)

We use Lemma~\ref{lem_sim} to study dissimilar batch decompositions of random datasets. Formally, a batch decomposition of the dataset~\(\{W_j\}\) is a set of~\(t\) mutually disjoint subsets~\({\cal V}_i \subset \{1,2,\ldots,n\}, 1 \leq i \leq t\) such that~\(\bigcup_{1 \leq i \leq t} {\cal V}_i = \{1,2,\ldots,n\}.\) We define~\({\cal V}_i\) to be the~\(i^{th}\) \emph{batch} and denote~\(t\) to be the \emph{size} of the decomposition.
\begin{definition}\label{def_dissim}
For integer~\(k \geq 1\) we say that~\(\{{\cal V}_j\}_{1 \leq j \leq t} \) is a~\(k-\)good batch decomposition if each~\({\cal V}_j\) contains at least one index of an  uncorrupted data point and every data point~\(W_v, v \in {\cal V}_j\) is similar to at most~\(k-1\) data points with indices in~\({\cal V}_j.\)
\end{definition}
In other words, any batch in a~\(k-\)good decomposition has at least one index of an uncorrupted data point and at most~\(k-1\) indices from the similarity set~\(N(v)\) for a data point~\(W_v\) whose index~\(v\) is present within the batch.

Letting~\(\tau_k\)  be the minimum size of a~\(k-\)good batch decomposition and recalling that~\(p_{up} = \max_{y \in {\cal Y}}p(y)\) is the maximum probability of occurrence of a categorical symbol, we have the following result.
\begin{theorem}\label{thm_one} Suppose the condition~(\ref{daldamma}) in Lemma~\ref{lem_sim} holds and let~\(\Delta,\Lambda\) and~\(\gamma_2 > 0\) be as in~(\ref{rasathi_ax}). Also suppose
\begin{equation}\label{zaveri}
\frac{\Lambda}{\log{n}} \longrightarrow \infty,\;\;\;(1-p_0)^{\varepsilon}\Delta \longrightarrow \infty\;\;\text{ and }\;\;\frac{\Delta \log{n}}{n(1-p_0)} \longrightarrow 0
\end{equation}
for some constant~\(\varepsilon > 0\) so that the bounds in~(\ref{rasathi_ax}) hold with high probability.\\
\((a)\) If either~\(p_0 =0\) and~\(1 \leq k \leq \gamma_2\Delta\) or~\(\beta \log{\Delta} \leq k \leq \gamma_2\Delta\) for some constant~\(\beta > 0,\) then there are constants~\(\lambda_1,\lambda_2 >0\) such that
\begin{equation}\label{rasathi}
\mathbb{P}\left(\frac{\lambda_1\Delta}{k} \leq \tau_k \leq \frac{\lambda_2\Delta}{k}\right) \geq 1-2ne^{-\lambda_1\Lambda}.
\end{equation}
\((b)\) If~\(p_0 =0\) so that~\(\Delta=  \Lambda = nr_n^dp_{up},\) then there are constants~\(\lambda_3,\lambda_4>  0\) such that for all~\(1 \leq k \leq \lambda_3 \log{n},\) we have
\begin{equation}\label{vaigai}
\mathbb{E}\left(\frac{\tau_k}{\mathbb{E}\tau_k}-1\right)^2 \leq \lambda_4 \cdot \frac{\log{n}}{\Delta}.
\end{equation}
\end{theorem}
The range of~\(k \leq \lambda_3\log{n}\) in part~\((b)\) is consistent with the range~\(k \leq \gamma_2\Delta\) in~\((a)\) since~\(p_0=0\) in part~\((b)\) and so~\(\Delta = \Lambda\) is much larger than~\(\log{n}\) by condition~(\ref{zaveri}).

From~(\ref{rasathi}) we see that the minimum size~\(\tau_k\) of a~\(k-\)good decomposition is of the order of~\(\frac{\Delta}{k}\) with high probability, i.e., with probability converging to one as~\(n \rightarrow \infty.\) As expected, relaxing the similarity constraint~\(k,\) i.e., increasing~\(k,\) results in smaller size decompositions and this underlines an inherent tradeoff between reducing batchwise similarity and the overall decomposition size.


\underline{\emph{Example}}: Suppose~\(p_0=0\)  and~\(p_{up} \geq \frac{c}{n^{\theta}}\) for some constants~\(\theta,c > 0,\) so that data points are uncorrupted and the highest probability of occurrence of a categorical element is at least of the order of~\(\frac{1}{n^{\theta}}.\) Setting~\(r_n = \frac{1}{n^{\beta}}\) for some~\(0 < \beta < \frac{1-\theta}{d}\) strictly, we then get from~(\ref{delta_def}) that
\begin{equation}\label{thodu_vanam}
\Lambda = \Delta = nr_n^{d} p_{up} \geq cn^{1-\theta-d\beta} \longrightarrow \infty.
\end{equation}
The conditions in~(\ref{zaveri}) are also satisfied and so~(\ref{rasathi}) provides  high probability bounds for the minimum size of a~\(k-\)good decomposition. In other words,  if we have knowledge of~\(\theta\) (which could also be  estimated from the dataset itself), then we can adjust~\(\beta\) accordingly to ensure that~(\ref{zaveri}) holds and thereby obtain near optimal batch decompositions in the sense of~(\ref{rasathi}).



For notational simplicity, we reuse constants~\(D_1,D_2,\ldots,\) etc. in proof below.\\
\emph{Proof of Theorem~\ref{thm_one}\((a)\)}:  The proof of lower deviation bound in~(\ref{rasathi}) is similar to that of~(\ref{rasathi_ax}) in Lemma~\ref{lem_sim}: Indeed, let~\(E_{tot}\) and~\(E_{cat}(y_0) \supset E_{tot}\) be the events as defined prior to~(\ref{e_tot_est}) in the proof of Lemma~\ref{lem_sim} and  let~\(\{{\cal V}_l\}_{1 \leq l \leq t}\) be any~\(k-\)good decomposition of minimum size~\(t = \tau_k.\) Since~\(E_{cat}(y_0)\) occurs, there is set~\({\cal S}\) containing at least~\(\frac{np_{up}p_0}{2}\) indices whose corresponding data points are corrupted and similar to each other. Any batch~\({\cal V}_l\) contains at most~\(k\) indices of~\({\cal S}\) and so~\(\tau_k \geq \frac{np_{up}p_0}{2k}.\) Similarly any two uncorrupted data points with continuous part in the square~\(S_1\) and having categorical part~\(y_0\) are similar to each other and since~\(E_{tot}\) occurs, there are at least~\(D_1nr_n^{d}p_{up}(1-p_0)\) such points for some constant~\(D_1 > 0.\) Again using the~\(k-\)good condition we must therefore have~\(t = \tau_{k} \geq \frac{D_1nr_n^{d}p_{up}(1-p_0)}{2k}.\) Combining the above two  bounds for~\(\tau_{k}\) and using~(\ref{e_tot_est}), we obtain the lower deviation bounds in~(\ref{rasathi}).

For obtaining the upper deviation bound in~(\ref{rasathi}), we see that the number~\(N_{unc}\) of uncorrupted data points is Binomially distributed with parameters~\(n\) and~\(1-p_0\) and defining~\(E_{unc} := \{N_{unc} \geq \frac{n(1-p_0)}{2}\}\) we get from~(\ref{conc_est_f}) that
\begin{equation}\label{e_unc_est}
\mathbb{P}\left(E_{unc}\right) \geq 1- \exp\left(-D_2n(1-p_0)\right) \geq 1-\exp\left(-D_3\Lambda\right)
\end{equation}
for some constants~\(D_2,D_3>  0,\) where the final bound in~(\ref{e_unc_est}) is true since~\(r_n\) is bounded by Theorem statement. Recalling the definition of the event~\(E_{sq}\) defined prior to~(\ref{sq_est}) in Lemma~\ref{lem_sim} and defining~\(E_{join} := E_{sq} \cap E_{unc},\) we get from~(\ref{sq_est}),~(\ref{e_unc_est})  and the union bound that
\begin{equation}\label{zilda}
\mathbb{P}\left(E_{join}\right) \geq 1- 2n\exp\left(-D_4\Lambda\right) - \exp\left(-D_3\Lambda\right)
\end{equation}
for some constant~\(D_4 > 0.\)

Assuming henceforth that~\(E_{join}\) occurs, we now perform a random assignment strategy as follows. We first consider the case~\(p_0 >0.\) Let~\(q = \frac{\theta \Delta}{k}\) for some constant~\(\theta > 0\) to be determined later and let~\((Z_1,\ldots,Z_n)\) be independent and identically distributed random variables in~\(\{1,2,\ldots,q\}\) (that are also independent of~\(\{W_i\}\)) with distribution~\(\mathbb{P}_Z.\) Assign index~\(i\) to batch~\(Z_i\) and let~\(\{{\cal U}_{l}\}_{1 \leq l \leq q}\) be the resulting batch decomposition. In what follows, we use the local lemma (and hence the probabilistic method) to show the \emph{existence}  of a good batch decomposition.

Since~\(E_{unc}\) occurs, there are at least~\(\frac{n(1-p_0)}{2}\) uncorrupted data points and so the~\(\mathbb{P}_Z-\)probability that  batch~\({\cal U}_l\) contains no uncorrupted data point is at most~\[\left(1-\frac{1}{q}\right)^{n(1-p_0)/2} \leq \exp\left(-\frac{n(1-p_0)}{2q}\right).\] Therefore if~\(F_{one}\) denotes the event that each batch contains at least one uncorrupted data point, we get from the union bound that
\begin{equation}\label{f_one_estal}
\mathbb{P}_Z(F_{one}) \geq 1- n\exp\left(-\frac{n(1-p_0)}{2q}\right) = 1-n\exp\left(-\frac{nk(1-p_0)}{2\theta \Delta}\right) \longrightarrow 1
\end{equation}
since~\(\frac{n(1-p_0)}{\Delta}\) is much larger than~\(\log{n},\) by~(\ref{zaveri}).

Next we use the local Lemma in Lemma~\ref{lemmax} to show that~\(\{{\cal U}_l\}\) is a~\(k-\)good decomposition with positive~\(\mathbb{P}_Z-\)probability. Indeed, if~\(A_v\) is the event that~\(k\) indices from the similarity set~\(N(v)\) (see~(\ref{n_sim_def})) are assigned to the same batch as~\(v,\) then defining~\(d = d(v) := \#N(v)\) and using the estimate~\({d \choose k} \leq \left(\frac{de}{k}\right)^{k},\) we have that
\begin{equation}\label{nimrat_tits}
\mathbb{P}_Z(A_v) \leq {d \choose k} \cdot \frac{1}{q^{k}} \leq \left(\frac{de}{kq}\right)^{k}.
\end{equation}
Because~\(E_{sq}\) occurs, we know that~\(d = d(v) \leq D_5\Delta\) for some constant~\(D_5 > 0\) and~\(q = \frac{\theta \Delta}{k}\) by choice. Plugging these into~(\ref{nimrat_tits}) we get
\begin{equation}\label{nimrat_tits2}
\mathbb{P}_{Z}(A_v) \leq \left(\frac{D_5e}{\theta}\right)^{k} \leq \left(\frac{D_5e}{\theta}\right)^{\beta \log{\Delta}},
\end{equation}
since~\(k \geq \beta \log{\Delta},\) by Theorem statement.

Let~\(L \geq 3\) be an integer to be determined later. If we choose~\(\theta >0\) large enough, then from~(\ref{nimrat_tits2}) we see that
\begin{equation}\label{nimrat_tits3}
\mathbb{P}_Z(A_v)  \leq \frac{1}{\Delta^L} =: \frac{z_v}{2}.
\end{equation}
Also the events~\(A_u\) and~\(A_v\) are dependent if and only if the corresponding similarity sets~\(N(u)\) and~\(N(v)\) share a common index. Since each~\(N(v)\) has size at most~\(D_5\Delta,\) we see that any~\(A_v\) therefore depends on at most~\((D_5\Delta)^2\) of the events in~\(\{A_w\}.\) Letting~\(u \sim v\) denote that~\(A_u\) is dependent on~\(A_v,\) we then get that
\begin{align}\label{zalpa}
z_v \prod_{u \sim v}(1-z_u) &= \frac{2}{\Delta^L} \left(1-\frac{2}{\Delta^L}\right)^{D_5^2\Delta^2}  \geq \frac{2}{\Delta^L}\left(1-\frac{2D_5^2\Delta^2}{\Delta^L}\right) \geq \frac{1}{\Delta^L} \geq \mathbb{P}_Z(A_v),
\end{align}
where the second inequality in~(\ref{zalpa}) is true since~\(\Delta \geq \Lambda \longrightarrow \infty\) by Theorem statement and the final estimate in~(\ref{zalpa}) follows from~(\ref{nimrat_tits3}). Thus the conditions in Lemma~\ref{lemmax}\((b)\) are satisfied and so letting~\(F_{two} := \bigcap_{1\leq v \leq n} A^c_v, \) we get that
\begin{equation}\label{f_two_estal}
\mathbb{P}_Z(F_{two}) \geq \prod_{v}(1-z_v) = \left(1-\frac{2}{\Delta^L}\right)^{n} \geq \exp\left(-\frac{4n}{\Delta^L}\right),
\end{equation}
since~\(1-x \geq e^{-2x}\) for all~\(x < \frac{1}{2}.\)

Combining~(\ref{f_one_estal}) and~(\ref{f_two_estal}) we get that
\begin{align}
\mathbb{P}_Z(F_{one} \cap F_{two}) &\geq \mathbb{P}_Z(F_{two}) - \mathbb{P}_Z(F^c_{one}) \nonumber\\
&\geq \exp\left(-\frac{4n}{\Delta^L}\right) - n\exp\left(-\frac{nk(1-p_0)}{2\theta \Delta}\right) \nonumber\\
&\geq e^{-I_1} - ne^{-I_2}  \nonumber\\
&= e^{-I_2}\left(e^{I_2-I_1} -n\right),\label{chillax}
\end{align}
where~\(I_1 := \frac{4n}{\Delta^{L}},I_2 := \frac{n(1-p_0)}{2\theta \Delta}\) and the third estimate in~(\ref{chillax}) is true since~\(k \geq 1.\) Setting~\(L = \max\left(3,1+\frac{1}{\varepsilon}\right)\)  and using the middle condition in~(\ref{zaveri}), we see that~\(I_1\) is much smaller than~\(I_2\) and so~\(I_2-I_1 > \frac{I_2}{2}\) for all~\(n\) large. From the final condition in~(\ref{zaveri}), we see that~\(I_2\) is much larger than~\(\log{n}\) and so~\(I_2 > 4\log{n}\) for all~\(n\) large.  The bound~(\ref{chillax}) thus implies that~\(F_{one} \cap F_{two}\) occurs with positive~\(\mathbb{P}_Z-\)probability.

Summarizing, if the joint event~\(E_{join}\) defined prior to~(\ref{zilda}) occurs, then there exists a~\(k-\)good batch decomposition of size~\(q = \frac{\theta \Delta}{k}\) and the estimate~(\ref{zilda}) therefore obtains the upper deviation bound in~(\ref{rasathi}) for the case~\(p_0 > 0.\)

For the case~\(p_0 =0\) and~\(1 \leq k \leq \gamma_2\Delta,\) we use a ``subset" version of the local Lemma. For a vertex~\(v\) let~\({\cal T}_k(v)\) be the set of all subsets of the similarity set~\(N(v),\) having size~\(k+1\) and set~\({\cal T}_k := \bigcup_{1 \leq v \leq n} {\cal T}_k(v).\) For~\({\cal C} \in {\cal T}_k,\) we let~\(A_{\cal C}\) be the event that all indices in~\({\cal C}\) are assigned to the same batch   (i.e.~\(Z_v = Z_u\) for any two data points~\(u,v \in {\cal C}\)) so that
\begin{equation}\label{eq_pa}
\mathbb{P}_Z(A_{\cal C}) = \frac{1}{q^{k}} =: \frac{y({\cal C})}{2}.
\end{equation}
Since the event~\(E_{sq}\) occurs each data point is similar to at most~\(D_3\Delta\) other data points and so for any~\(v,\) there are at most~\({D_3\Delta \choose k} \leq \left(\frac{D_3\Delta e}{k}\right)^{k} =: L\) subsets in~\({\cal T}_k(v).\) By definition~\({\cal C}\) contains~\(k+1\) indices and so the event~\(A_{\cal C}\) is dependent on at most\\\((k+1) \cdot L \leq 2kL\) of the events in set~\(\{A_{{\cal D}}\}_{{\cal D} \in {\cal T}_k}.\)

Using the notation~\({\cal C} \sim {\cal D}\) to denote that~\({\cal C}\) and~\({\cal D}\) share a common data point, we then get that
\begin{eqnarray}
y({\cal C}) \prod_{{\cal D} \sim {\cal C}} (1-y({\cal D})) &\geq& \frac{2}{q^{k}} \left(1-\frac{2}{q^{k}}\right)^{2kL} \nonumber\\
&\geq& \frac{2}{q^{k}} \left(1-\frac{2kL}{q^{k}}\right) \nonumber\\
&=& \frac{2}{q^{k}} \left(1-2k\left(\frac{D_3e}{\theta}\right)^{k}\right) \nonumber\\
&\geq& \frac{1}{q^{k}} \nonumber\\
&=& \mathbb{P}_Y(A_{\cal C}) \nonumber
\end{eqnarray}
provided~\(\left(\frac{D_3e}{\theta}\right)^{k} \leq \frac{1}{4k} \) or equivalently if~\(\theta \geq (4k)^{1/k} \cdot D_3e\) for all~\(k \geq 1.\) Using~\(k^{1/k} \leq 4\) for all~\(k \geq 1\) we see that it suffices to ensure that~\(\theta \geq 8D_3e.\) Fixing such a~\(\theta\) we then get from Lemma~\ref{lemmax}\((ii)\)  that with positive~\(\mathbb{P}_Z-\)probability, there is a~\(k-\)good decomposition~\(\{{\cal V}_l\}_{1 \leq l \leq q}.\) This obtains the upper deviation bound for~\(\tau_k\) for the case~\(p_0~=~0\) and therefore completes the proof of Theorem~\ref{thm_one}\((a).\)~\(\qed\)


\emph{Proof of Theorem~\ref{thm_one}\((b)\)}: To obtain the variance bound for~\(\tau_k,\) we use the martingale difference method. Let~\(z \geq 1\) be an integer and suppose for simplicity that~\(\frac{n}{z}\) is an integer. For~\(1 \leq j \leq \frac{n}{z},\) we let~\({\cal F}_j\) be the sigma-field generated by the data points~\(\{W_u\}_{1 \leq u \leq jz}\) and get from the martingale difference property that
\begin{equation}\label{mart_diff}
var(\tau_k) = \sum_{j=1}^{n/z}\mathbb{E}\left(\mathbb{E}(\tau_k \mid {\cal F}_{j}) -\mathbb{E}(\tau_k \mid {\cal F}_{j-1}) \right)^2.
\end{equation}
We rewrite the right side of~(\ref{mart_diff}) in a more convenient form as follows. For~\(1 \leq j \leq \frac{n}{z},\)  suppose we replace the~\(z\) data points~\(\{W_{(j-1)z+1},\ldots,W_{jz}\}\) with independent copies~\(\{W^{(c)}_{(j-1)z+1},\ldots,W^{(c)}_{jz}\}\) that are also independent of all random variables defined so far. We define~\(\tau_k^{(j)}\) to be the minimum size of a~\(k-\)good batch decomposition  of the modified dataset
\begin{equation}\label{mod_data}
\left(\{W_l\}_{1 \leq l  \leq n} \setminus \{W_{(j-1)z+1},\ldots,W_{jz}\} \right) \bigcup \{W^{(c)}_{(j-1)z+1},\ldots,W^{(c)}_{jz}\}.
\end{equation}
With the above notations, we have
\begin{align}
\left(\mathbb{E}(\tau_k \mid {\cal F}_{j}) -\mathbb{E}(\tau_k \mid {\cal F}_{j-1}) \right)^2 &= \left(\mathbb{E}\left(\tau_k-\tau_k^{(j)} \mid {\cal F}_{j-1}\right)\right)^2 \nonumber\\
&\leq \mathbb{E}\left(\left(\tau_k-\tau_k^{(j)}\right)^2 \mid {\cal F}_{j-1}\right)
\label{nikki_tits2}
\end{align}
by the Jensen's inequality for conditional expectations. Taking expectations on both sides of~(\ref{nikki_tits2}) and plugging into~(\ref{mart_diff}), we get
\begin{equation}\label{mart_diff2}
var(\tau_k) \leq \sum_{j=1}^{n/z}\mathbb{E}\left(\tau_k-\tau_k^{(j)}\right)^2.
\end{equation}

By definition, the minimum size~\(\tau_k = \tau_k(n)\) of a~\(k-\)good decomposition is non-decreasing in the size of the dataset~\(n\) and so if~\({\cal V}_{1},\ldots,{\cal V}_t, t =: \tau_k(j;n-z)\) is a minimum size~\(k-\)good batch decomposition of the data subset~\(\{W_l\}_{1 \leq l  \leq n} \setminus \{W_{(j-1)z+1},\ldots,W_{jz}\}\) containing~\(n-z\) data points, then~\(\tau_k(j;n-z) \leq \tau_k(n).\) Next if~\({\cal U}_1,\ldots,{\cal U}_m ,m =: \tau_k(j;z)\) is a minimum size~\(k-\)good batch decomposition of~\(\{W_{(j-1)z+1},\ldots,W_{jz}\}\) then the~\(t+m\) batches~\({\cal V}_j, 1 \leq j \leq t\) and~\({\cal U}_{l}, 1 \leq l \leq m\) together form a~\(k-\)good batch decomposition of the overall dataset~\(\{W_i\}_{1 \leq i \leq n}.\)

From the discussion in the previous paragraph we therefore see that
\begin{equation}\label{ondra}
\tau_{k}(j;n-z) \leq \tau_{k}(n) \leq \tau_{k}(j;n-z) + \tau_k(j;z),
\end{equation}
where~\(\tau_k(j;z)\) has the same distribution as~\(\tau_k(z).\) Similarly, if~\(\tau^{(c)}_k(j;z)\) is the minimum size of a~\(k-\)good batch decomposition of the copy~\(\{W^{(c)}_{(j-1)z+1},\ldots,W^{(c)}_{jz}\},\) then from~(\ref{ondra}), we get that
\begin{equation}\label{renda}
\tau_{k}(j;n-z) \leq \tau^{(j)}_{k}(n) \leq \tau_{k}(j;n-z) + \tau^{(c)}_k(j;z),
\end{equation}
where we recall from the description prior to~(\ref{mart_diff2}) that~\(\tau^{(j)}_k = \tau^{(j)}_k(n)\) is the minimum size of a~\(k-\)size decomposition of the modified dataset in~(\ref{mod_data}). Again~\(\tau^{(c)}_k(j;z)\) is identically distributed as~\(\tau_k(z).\)

Combining~(\ref{ondra}) and~(\ref{renda}), we see that~\[\left|\tau_k(n) - \tau^{(j)}_{k}(n)\right| \leq \tau_k(j;z) + \tau_k^{(c)}(j;z)\]
and so squaring and taking expectations we get
\begin{align}\label{n_free_diff2}
\mathbb{E}\left(\tau_k(n) - \tau^{(j)}_{k}(n)\right)^2 &\leq \mathbb{E}\left(\tau_k(j;z) + \tau_k^{(c)}(j;z)\right)^2 \nonumber\\
&\leq 2\mathbb{E}\tau^{2}_k(j;z) + 2\mathbb{E}\left(\tau_k^{(c)}(j;z)\right)^2\nonumber\\
&= 4\mathbb{E}\tau^2_{k}(z)
\end{align}
by symmetry.

Plugging~(\ref{n_free_diff2}) into~(\ref{mart_diff2}) we get that
\begin{equation}\label{ayan}
var(\tau_k(n)) \leq 4\sum_{j=1}^{n/z}\mathbb{E}\tau_k^2(z) = \frac{4n}{z} \mathbb{E}\tau^{2}_k(z).
\end{equation}
To estimate~\(\mathbb{E}\tau_k(z),\) we use~(\ref{rasathi}) with~\(p_0 = 0\) and~\(n\) replaced by~\(z.\)  In this case~\(\Delta = \Lambda = nr_n^{d}p_{up}\) and recalling the constant~\(\gamma_2\) in Theorem statement, we then get for all~\(1 \leq k \leq \gamma_2\Delta(z)\)  that~\[\mathbb{P}\left(\tau_k(z) \leq \frac{\lambda_2}{k}\Delta(z)\right) \geq 1-2ze^{-\lambda_1\Delta(z)},\]
where~\(\Delta(z) := zr_n^{d}p_{up} = \frac{z\Delta}{n}\) and~\(\lambda_1,\lambda_2 > 0\) are the constants in~(\ref{rasathi}). Thus using the bound~\(\tau_k(z) \leq z\) we get that
\begin{equation}\label{bheema}
\mathbb{E}\tau_k^2(z) \leq \frac{\lambda_2^2}{k^2}\Delta^2(z) + z^2 \cdot 2z e^{-\lambda_1 \Delta(z)}.
\end{equation}

We now choose~\(z\) appropriately so that the second term on the right side of~(\ref{bheema}) is small compared to the first term. Specifically, we set~\(z := \frac{6 n\log{n}}{\lambda_1\Delta}\) so that\\\(\frac{z}{n} =\frac{6}{\lambda_1} \cdot \frac{\log{n}}{\Delta} \longrightarrow 0\) by Theorem statement. For all~\(n\) large, our choice of~\(z\) is therefore valid and so~\[z \leq n,\;\;\; e^{-\lambda_1\Delta(z)} = \frac{1}{n^{6}} \text{ and }\Delta(z) = \frac{z\Delta}{n} = \frac{6\log{n}}{\lambda_1}\] for all~\(n\) large. From~(\ref{bheema}), we therefore get for all~\(1 \leq k \leq \frac{6\gamma_2 \log{n}}{\lambda_1}\) that
\begin{equation}\label{tella}
\mathbb{E}\tau_{k}^2(z) \leq \frac{\lambda_2^2}{k^2}\Delta^2(z) + \frac{2n^3}{n^{6}} = \lambda_2^2 \frac{z^2\Delta^2}{k^2n^2} + \frac{2}{n^{3}}.
\end{equation}

Since~\(r_n\) is bounded by Theorem statement we have that~\[z =\frac{6\log{n}}{\lambda_1p_{up}r_n^{d}} \geq D_1 \log{n}\] for some constant~\(D_1 > 0\) and moreover from the condition~(\ref{zaveri}), we see that~\(\Delta = \Lambda\) is much larger than~\(\log{n}.\) Therefore for all~\(k \leq \frac{6\gamma_2\log{n}}{\lambda_1} \leq \Delta,\) we get that~\[\lambda_2^2\frac{z^2\Delta^2}{k^2n^2} \geq \frac{\lambda_2^2z^2}{n^2} \geq D_2 \cdot \left(\frac{\log{n}}{n}\right)^2 \geq \frac{2}{n^3}\] for all~\(n\) large and some constant~\(D_2 >0.\) From~(\ref{tella}) we therefore get that\\\(\mathbb{E}\tau_k^2(z)\leq 2\lambda_2\frac{z^2\Delta^2}{k^2n^2}\) and plugging this into~(\ref{ayan}), we get for all~\(1 \leq k \leq \frac{6\gamma_2\log{n}}{\lambda_1}\) that
\begin{equation}\label{roja_kadale}
var(\tau_k) \leq \frac{4n}{z} \cdot 2\lambda_2 \frac{z^2\Delta^2}{k^2n^2} = D_3 \frac{\log{n}}{\Delta} \cdot \frac{\Delta^2}{k^2}
\end{equation}
for some constant~\(D_3 > 0.\) Finally, the lower bound in~(\ref{rasathi}) together with the fact that~\(\frac{\Delta}{\log{n}} = \frac{\Lambda}{\log{n}} \longrightarrow \infty\) (see~(\ref{zaveri})),  implies that~\(\mathbb{E}\tau_k \geq \frac{D_4\Delta}{k}\) for some constant~\(D_4 > 0.\) Substituting this into~(\ref{roja_kadale}), we get the variance estimate in~(\ref{rasathi}) and this completes the proof of the Theorem.~\(\qed\)

In our final result, we study the \emph{maximum} size of data subsets  with a given similarity. For analytical convenience, we assume henceforth that the corruption probability~\(p_0=0\) and our analysis below holds also for~\(p_0 > 0.\) For a subset~\({\cal V} \subset \{1,2,\ldots,n\}\) and integer~\(k \geq 1,\) we say that~\(\{W_v\}_{v \in {\cal V}}\) has similarity at most~\(k-1\) if each~\(W_v, v \in {\cal V}\) is similar to at most~\(k-1\) other points~\(W_u, u \in {\cal V}.\) A subset having similarity zero is also called a  \emph{similarity-free} subset.

Letting~\(N_{sim}(k)\) be the maximum size of a subset of the dataset~\(\{W_j\}_{1 \leq j \leq n}\) having similarity at most~\(k-1,\) we are interested to see how~\(N_{sim}(k)\) varies with the similarity constraint~\(k\)  and also how the continuous and categorical parts of the data affect~\(N_{sim}(k).\) Intuitively, we expect~\(N_{sim}(k)\) to increase with~\(k\) and  in fact if condition~(\ref{daldamma}) of Theorem~\ref{thm_one} holds, then the proof of Theorem~\ref{thm_one}  implies with probability at least~\(1-ne^{-\theta_1 nr_n^{d}p_{up}}\) there exists a~\(k-\)good batch decomposition~\(\{{\cal V}_l\}_{1 \leq l \leq t}\) of size~\(t \leq \frac{\theta_2nr_n^{d}p_{up}}{k}\) for some constants~\(\theta_1,\theta_2 > 0.\)

Since there are~\(n\) data points in total, the pigeonhole principle asserts that there necessarily exists a set~\({\cal V}_{l_0}\subset\{1,2,\ldots,n\}\) of size~\(\#{\cal V}_{l_0} \geq \frac{k}{\theta_2 r_n^{d}p_{up}}.\) By definition, the set of data points with indices in~\({\cal V}_{l_0}\) have similarity at most~\(k-1\) and so
\begin{equation}\label{emma_tits}
\mathbb{P}\left(N_{sim}(k) \geq \frac{k}{\theta_2r_n^{d}p_{up}}\right) \geq 1-ne^{-\theta_1nr_n^{d}p_{up}}.
\end{equation}

In the following result, we obtain stronger bounds for~\(N_{sim}(k)\) in terms of the size of the categorical space~\({\cal Y}.\) As before, constants do not depend on~\(n\) and we recall that~\(S_f\) is the support of density~\(f\) of the continuous part of the data point and~\(p(.)\) is the distribution of the categorical part. We also use the notation~\(k=o(n)\) to denote that~\(\frac{k}{n} \longrightarrow 0\) as~\(n \rightarrow \infty.\)
\begin{theorem}\label{vanilla_case} Suppose the corruption probability~\(p_0=0\)  and also suppose that conditions~(\ref{daldamma}) in Lemma~\ref{lem_sim} and~(\ref{zaveri}) in Theorem~\ref{thm_one} holds. For any integer~\(k = k(n) \geq 1,\) the variance~\(var(N_{sim}(k)) \leq 4\mathbb{E}N_{sim}(k).\) Moreover, there are constants~\(\beta_1,\beta_2> 0\) such that if~\(k = o(n),\) then
\begin{equation}\label{n_free_low_two}
\mathbb{P}\left(N_{sim}(k) \geq \frac{\beta_1 k\#{\cal Y}}{r_n^d}\cdot (1-\zeta)\right) \geq 1-\frac{\beta_2 k r_n^d}{\#{\cal Y}(1-\zeta)},
\end{equation}
where
\begin{equation}\label{zeta_def}
\zeta := \frac{1}{\#{\cal Y}}\sum_{y \in {\cal Y}} \exp\left(-\epsilon_{up}\pi_d \frac{nr_n^{d} p(y)}{k}\right),
\end{equation}
~\(\epsilon_{up}\) is the density upper bound in~(\ref{daldamma}) and~\(\pi_d\) is the volume of the unit ball in~\(d-\)dimensions. Conversely, if the support~\(S_f\) has constant side length, then
\begin{equation}\label{virat_50}
N_{sim}(k) \leq \frac{C k \#{\cal Y}}{r_n^d}
\end{equation}
for some constant~\(C > 0.\)
\end{theorem}
As illustration we continue with the example described after Theorem~\ref{thm_one} (see discussion containing~(\ref{thodu_vanam})).\\
\emph{\underline{Example (contd)}}: If~\(p_{low} := \min_{y \in {\cal Y}}p(y)\)  is the minimum probability of occurrence of a categorical element, then using~\(\sum_{y\in{\cal Y}} p(y) = 1\) we immediately get that
\begin{equation}\label{gilma}
p_{low} \leq \frac{1}{\#{\cal Y}} \leq p_{up} = \max_{y \in {\cal Y}}p(y),
\end{equation}
the maximum probability of occurrence of a categorical element.

Suppose that
\begin{equation}\label{cat_assum}
\#{\cal Y} = n^{\rho},\;\;p_{up} = \frac{c}{n^{\theta}} \text{ and } p_{low} = \frac{1}{n^{\lambda}}
\end{equation}
for some constants~\(c> 0 \) and~\(0 < \rho,\theta,\lambda <1.\) From~(\ref{gilma}), we see that~\(\lambda \geq \rho > \theta\) necessarily and so if we set~\(r_n = \frac{1}{n^{\beta}}\) with
\begin{equation}\label{beta_choice}
0 < \beta  <\frac{1-\lambda}{d} < \frac{1-\theta}{d}
\end{equation}
strictly, then by the argument in the paragraph containing~(\ref{thodu_vanam}), we see that the decomposition size bounds in~(\ref{rasathi}) of Theorem~\ref{thm_one} hold with high probability for all~\(1 \leq k \leq \gamma_2\Delta,\) where~\(\gamma_2\) is the constant in~(\ref{rasathi_ax}). This in turn implies that the lower bound in~(\ref{emma_tits}) holds with high probability.

To compare the bounds in Theorem~\ref{vanilla_case} with~(\ref{emma_tits}), we first get from~(\ref{thodu_vanam}) that~\[\Delta =nr_n^dp_{up} = cn^{1-d\beta-\theta} \longrightarrow \infty,\] by our choice of~\(\beta\) in~(\ref{beta_choice}). Therefore we consider~\(k = k(n) = n^{\delta},\) where\\\(0 \leq \delta < 1-d\beta-\theta\) is chosen so that
\begin{equation}\label{t_cond}
\frac{nr_n^d p_{low}}{k} \longrightarrow \infty\;\;\;\;\text{ and }\;\;\;\; \frac{k \cdot r_n^d}{\#{\cal Y}} \longrightarrow 0
\end{equation}
or equivalently that
\[1-d\beta - \lambda - \delta > 0 \text{ and } \delta  < \rho+d\beta .\]

For all~\[0 \leq \delta < \delta_0 := \min\left(1-d\beta-\lambda,1-d\beta-\theta,\rho+d\beta\right)\] (which is positive, again by our choice of~\(\beta\) in~(\ref{beta_choice})), we get from~(\ref{t_cond}) and the definition of~\(\zeta\) in~(\ref{zeta_def})  that~\[\zeta \leq \exp\left(-\epsilon_{up} \pi_d \frac{nr_n^{d}p_{low}}{k}\right)\longrightarrow 0\] as~\(n \rightarrow \infty.\) By the upper bound~(\ref{virat_50})  and the lower bound~(\ref{n_free_low_two}), we then deduce that there are constants~\(C_1,C_2 > 0\) such that
\begin{equation}\label{sydney_tits}
\mathbb{P}\left( \frac{C_1 k \#{\cal Y}}{r_n^d} \leq N_{sim}(k) \leq \frac{C_2 k\#{\cal Y}}{r_n^d}\right) \longrightarrow 1
\end{equation}
as~\(n \rightarrow \infty.\) In other words~\(N_{sim}(k)\) is of the order of~\(\frac{k \cdot \#{\cal Y}}{r_n^d}\) and therefore linear in the similarity constraint and the size of the categorical space, with high probability. But from~(\ref{cat_assum}) and the fact that~\(\rho > \theta\) strictly, we get that~\(\frac{1}{p_{up}}\) is much smaller than~\(\#{\cal Y}\) and consequently, the lower bound in~(\ref{sydney_tits}) is much larger than~(\ref{emma_tits}).



\emph{Proof of Theorem~\ref{vanilla_case}}: We use the martingale difference method to obtain  the variance bound for~\(N_{sim} = N_{sim}(t).\) For~\(1 \leq j \leq n,\) suppose we replace the data point~\(W_j\) with an independent copy~\(W^{(c)}_j\) that is also independent of all random variables defined so far. We define~\(N_{sim}^{(j)}\) to be the maximum size of a subset of the modified dataset~\(\{W_l\}_{1 \leq l \neq j \leq n} \cup \{W^{(c)}_j\},\) having similarity at most~\(k-1.\) With these notations, we have from Efron-Stein inequality (see~\cite{steele}) that
\begin{equation}\label{mart_diff2_red}
var(N_{sim}) \leq \sum_{j=1}^{n}\mathbb{E}\left(N_{sim}-N_{sim}^{(j)}\right)^2.
\end{equation}

Let~\({\cal S}_{sim}\) be a subset of maximum size~\(N_{sim}\) in the dataset~\(\{W_i\}\) having similarity at most~\(k-1\) and suppose we replace the data point~\(W_j\) with an independent copy~\(W_{j}^{(c)}.\) If~\({\cal S}^{(j)}_{sim}\) is a subset of maximum size~\(N_{sim}^{(j)}\) having similarity at most~\(k-1\) in the modified dataset~\(\{W_l\}_{l \neq j} \cup \{W^{(c)}\}_j,\) then we must have~\(\left|N_{sim}-N_{sim}^{(j)}\right| \leq 1\) and moreover,~\(N_{sim}^{(j)} \neq N_{sim}\) only if either~\(W_j \in {\cal S}_{sim}\) or~\(W^{(c)}_j \in {\cal S}^{(j)}_{sim}.\) Thus letting~\(\ind(.)\) denote the indicator function, we have that
\begin{eqnarray}\nonumber
\left|N_{sim} - N^{(j)}_{sim}\right| &\leq& \ind\left(\left\{W_j \in {\cal S}_{sim}\right\} \cup \left\{W^{(c)}_j \in {\cal S}^{(c)}_{sim}\right\}\right) \nonumber\\
&\leq& \ind\left(W_j \in {\cal S}_{sim}\right) + \ind\left(W^{(c)}_j \in {\cal S}^{(c)}_{sim}\right)
\end{eqnarray}
and squaring and taking expectations we get
\begin{eqnarray}\label{n_free_diff2_red}
\mathbb{E}\left(N_{sim} - N^{(j)}_{sim}\right)^2 &\leq& \mathbb{E}\left(\ind\left(W_j \in {\cal S}_{sim}\right) + \ind\left(W^{(c)}_j \in {\cal S}^{(c)}_{sim}\right)\right)^2 \nonumber\\
&\leq&  2\mathbb{E}\ind\left(W_j \in {\cal S}_{sim}\right) + 2\mathbb{E}\ind\left(W^{(c)}_j \in {\cal S}^{(c)}_{sim}\right) \nonumber\\
&=& 4\mathbb{P}\left(W_j \in {\cal S}_{sim}\right),
\end{eqnarray}
by symmetry.

Plugging~(\ref{n_free_diff2_red}) into~(\ref{mart_diff2_red}) we get that
\[var(N_{sim}) \leq 4\sum_{j=1}^{n}\mathbb{P}\left(W_j \in {\cal S}_{sim}\right) = 4\mathbb{E}\sum_{j=1}^{n} \ind\left(W_j \in {\cal S}_{sim}\right) = 4\mathbb{E}N_{sim}\] and this obtains the desired variance bound for~\(N_{sim} = N_{sim}(k).\)

To get the upper bound for~\(N_{sim}(k),\) we assume for simplicity that the support~\(S_f\) is the square of unit side length centred at the origin and first obtain the bound for the case~\(k=1.\) We divide~\(S_f\) into disjoint~\(\frac{r_n}{\sqrt{4d}} \times \frac{r_n}{\sqrt{4d}}\) squares~\(\{R_j\}_{1 \leq j \leq N} \) where we again assume for simplicity that~\(N = \left(\frac{\sqrt{4d}}{r_n}\right)^{d}\) is an integer.

We recall that~\(X_l\) and~\(X_k\) denote the continuous parts of the~\(l^{th}\) and~\(k^{th}\) data points, respectively. If both~\(X_l\) and~\(X_k\) belong to~\(R_j\) then by our choice of the side length of~\(R_j,\) we see that~\(d(X_l,X_k) < r_n.\) Thus if~\({\cal J}\) is any similarity-free subset of the dataset~\(\{W_i\}_{1 \leq i \leq n},\) then there are at most~\(\#{\cal Y}\) data points of~\({\cal J}\) present in any~\(R_j\) and so the size of~\({\cal J}\) is at most~\(N \cdot \#{\cal Y};\)
i.e.,
\begin{equation}\label{n_red_one_ax}
N_{sim}(1) \leq N \cdot \#{\cal Y} = \left(\frac{\sqrt{4d}}{r_n}\right)^{d} \cdot \#{\cal Y}.
\end{equation}
This proves~(\ref{virat_50}) for the case~\(k=1.\)

The proof for general case in analogous. If~\({\cal T}\) is a subset of the dataset~\(\{W_i\}_{1 \leq i \leq n}\) having similarity at most~\(k-1,\) then for each~\(R_j\)  and each~\(y \in {\cal Y}\) there are at most~\(k\) data points of~\({\cal T}\) present in~\(R_j.\) Thus any~\(R_j\) contains at most~\(k \cdot \#{\cal Y}\) data points of~\({\cal T}\) and arguing as before, we get~(\ref{virat_50}).

Finally, for the lower bound~(\ref{n_free_low_two}), we use iteration as follows. Setting~\(k=1\) in the variance estimate for~\(N_{free} := N_{sim}(1)\) and using the Chebychev's inequality, we get for~\(\epsilon > 0\) that
\[\mathbb{P}\left(N_{free} \geq (1-\epsilon)\mathbb{E}N_{free}\right) \geq 1-\frac{1}{\epsilon^2} \cdot var\left(\frac{N_{free}}{\mathbb{E}N_{free}}\right) \geq 1-\frac{4}{\epsilon^2\mathbb{E}N_{free}}.\] Fixing~\(\epsilon = \frac{1}{2}\) we therefore have
\begin{equation}\label{n_free_ha}
\mathbb{P}\left(N_{free} \geq \frac{\mathbb{E}N_{free}}{2}\right) \geq 1-\frac{16}{\mathbb{E}N_{free}}.
\end{equation}

We now obtain a lower bound for~\(\mathbb{E}N_{free}\) by an iterative procedure of adding the data points one by one. If~\(N_{free}(n)\) is the maximum size of a redundancy-free set corresponding to the dataset~\(\{W_j\}_{1 \leq j \leq n},\) then we see that
\[N_{free}(n) \geq N_{free}(n-1) + \ind(A_n),\]
where~\(A_n\) is the event that for each~\(1 \leq j \leq n-1\) we either have~\(Y_n \neq Y_j\) or~\(d(X_n,X_j) > r_n.\) Thus
\begin{equation}\label{e_n_free}
\mathbb{E}N_{free}(n) \geq \mathbb{E}N_{free}(n-1) + \mathbb{P}(A_n)
\end{equation}
and  using Fubini's theorem we have
\begin{equation}\label{an_exp}
\mathbb{P}(A_n) = \sum_{y}\int \mathbb{P}\left(A_n(x,y)\right) p(y) f(x) dx
\end{equation}
where
\begin{equation}\label{an_xy}
A_n(x,y) := \bigcap_{j=1}^{n-1} \{Y_j \neq y\} \bigcup \left\{d(x,X_j) > r_n\right\}
\end{equation}

To estimate~\(\mathbb{P}(A_n(x,y)),\) we let~\(B(x,r_n)\) denote the ball with centre~\(x\) and radius~\(r_n\) to get
\begin{equation}\label{dudu}
\mathbb{P}\left(d(x,X_j) < r_n \right) = \int_{B(x,r_n)} f(z)dz  \leq  \epsilon_{up} \pi_d r_n^{d}
\end{equation}
by the upper bound~(\ref{daldamma}) for the density. By the independence of the categorical and continuous parts, we then see that
\begin{equation}\label{dodo}
\mathbb{P}\left(\left\{d(x,X_j) < r_n\right\} \bigcap \left\{Y_j=y\right\}\right) = \mathbb{P}(d(x,X_j) < r_n) \mathbb{P}(Y_j=y) \leq \theta(y),
\end{equation}
where~\(\theta(y) := \epsilon_{up} \pi_d r_n^{d} p(y).\) From~(\ref{dodo}) and the fact that the data points are independent, we obtain~\[\mathbb{P}(A_n(x,y)) \geq \left(1-\theta(y)\right)^{n-1}\] and plugging this into~(\ref{an_exp}) we finally get
\begin{equation}\label{dolli}
\mathbb{P}(A_n) \geq \sum_{y} \left(1-\theta(y)\right)^{n-1}p(y).
\end{equation}

Substituting~(\ref{dolli}) into~(\ref{e_n_free}), we get that
\[\mathbb{E}N_{free}(n) \geq \mathbb{E}N_{free}(n-1) + \sum_{y} \left(1-\theta(y)\right)^{n-1} p(y)\] and proceeding iteratively, we have
\begin{eqnarray}\label{jolly}
\mathbb{E}N_{free}(n) &\geq& \sum_{k=0}^{n-1} \left(1-\theta(y)\right)^{k}p(y) \nonumber\\
&=& \sum_{y} \frac{p(y)}{\theta(y)} \cdot \left(1-(1-\theta(y))^{n}\right) \nonumber\\
&\geq& \sum_{y}\frac{p(y)}{\theta(y)} \left(1-e^{-n\theta(y)}\right) \nonumber\\
&=& \frac{\#{\cal Y}}{\epsilon_{up}\pi_d r_n^d}\left(1-\zeta\right),
\end{eqnarray}
where~\(\zeta\) is as defined in~(\ref{n_free_low_two}) with~\(t=1.\) Plugging~(\ref{jolly}) into~(\ref{n_free_ha}), we obtain~(\ref{n_free_low_two}) for the case~\(k=1.\)

For general~\(k,\) we split the dataset~\(\{W_i\}_{1 \leq i \leq n}\) into~\(k\) disjoint subsets~\({\cal I}_{j}, 1 \leq j \leq k\) each of size~\(\frac{n}{k},\) where we assume for simplicity that~\(\frac{n}{k}\) is an integer (else we simply throw away at most~\(k = o(n)\) data points from~\(\{W_i\}\) so that the size of the remaining set is a multiple of~\(k\)). If~\({\cal S}_{free}(j)\) is a similarity-free set in~\({\cal I}_j\) of maximum size, then from the basis step above, we get for each~\(1 \leq j \leq k\) that
\begin{equation}\label{divy}
\mathbb{P}\left(\#{\cal S}_{free}(j) \geq \frac{\beta_1}{r_n^{d}}\#{\cal Y} (1-\zeta)\right) \geq 1-\frac{\beta_2r_n^{d}}{\#{\cal Y} (1-\zeta)}
\end{equation}
where~\(\zeta\) is as defined in~(\ref{n_free_low_two}). By construction, the union~\(\bigcup_{1 \leq  j \leq k}\{{\cal S}_{free}(j)\}\) has similarity at most~\(k-1\) and so applying the union bound on~(\ref{divy}), we then get~(\ref{n_free_low_two}) for general~\(k.\) This completes the proof of Theorem~\ref{vanilla_case}.~\(\qed\)

\subsection*{\em Data Availability Statement}
Data sharing not applicable to this article as no datasets were generated or analysed during the current study.

\subsection*{\em Acknowledgement}
I thank Professors Rahul Roy, Federico Camia, Alberto Gandolfi, C. R. Subramanian and the referees for crucial comments that led to an improvement of the paper. We also thank IMSc and IISER Bhopal for my fellowships.

\subsection*{\em Conflict of Interest and Funding Statement}
I certify that there is no actual or potential conflict of interest in relation to this article. No funds, grants or other support was received for the preparation of this manuscript.


\begin{thebibliography}{99}
\bibitem{alon} N. Alon and J. Spencer. (2008).
\newblock{\em The Probabilistic Method}.
\newblock{Wiley Interscience}.


\bibitem{fern} A. Fern\'andez, S. del R\'io, N. V. Chawla and F. Herrera. (2017).
\newblock{An Insight into Imbalanced Big Data Classification: Outcomes and Challenges}.
\newblock{\em Complex Intelligent Systems}, \textbf{3}, 105.

\bibitem{ganesan} G. Ganesan. (2023).
\newblock{Probabilistic Bounds for Data Storage With Feature Selection and Undersampling}.
\newblock{\em Accepted for publication in Mathematical Sciences for Advancement of Science and Technology, (MSAST) 2023}.
\newblock{arxiv Link: https://arxiv.org/pdf/2301.04808}.

\bibitem{gupta} P. Gupta and P. R. Kumar. (1998).
\newblock {Critical Power for Asymptotic Connectivity in Wireless Networks}.
\newblock {\em Stochastic Analysis, Control, Optimization and Applications}, pp. 2203--2214.


\bibitem{kuhn} M. Kuhn  and K. Johnson. (2013).
\newblock{\em Applied Predictive Modeling}.
\newblock{Springer}.

\bibitem{mike} Y. He, G. Zhang and C-H. Hsu. (2021).
\newblock{\em Multiple Imputation of Missing Data in Practice}.
\newblock{CRC Press}.

\bibitem{penrose} M. Penrose. (2003).
\newblock {\em Random Geometric Graphs}.
\newblock {Oxford University Press}.


\bibitem{sharbaf} F. V. Sharbaf, S. Mosafer, M. H. Moattar. (2016).
\newblock{A Hybrid Gene Selection Approach for Microarray Data Classification using Cellular Learning Automata and Ant Colony Optimization}.
\newblock{\em Genomics}, \textbf{107}, pp. 231--238.

\bibitem{steele} J. M. Steele. (1986).
\newblock{An Efron-Stein Inequality for Nonsymmetric Statistics}.
\newblock{\em The Annals of Statistics}, \textbf{14}, pp. 753--758.

\bibitem{sui} Y. Sui, X. Zhang, J. Huan and H. Hong. (2019).
\newblock{ Exploring Data Sampling Techniques for Imbalanced Classification Problems}.
\newblock{\em  Proceedings SPIE 11198, Fourth International Workshop on Pattern Recognition, 1119813 (31 July 2019)}.

\bibitem{xu} M. Xu, S. Yoon, A. Fuentes and D. S. Park. (2023).
\newblock{A Comprehensive Survey of Image Augmentation Techniques for Deep Learning}.
\newblock{\em Pattern Recognition}, \textbf{137}, 109347.

\bibitem{zawbaa} H. M. Zawbaa, E. Emary, C. Grosan and V. Sansel. (2018).
\newblock{Large-dimensionality Small-instance Set Feature Selection: A Hybrid Bio-inspired Heuristic Approach}.
\newblock{\em Swarm and Evolutionary Computation}, \textbf{42}, pp. 29--42.


\end{thebibliography}
\end{document}